\documentclass[10pt,english]{article}
\usepackage{subfigure}
\usepackage{graphicx}
\usepackage{float}
\usepackage[T1]{fontenc}
\usepackage[latin9]{inputenc}
\usepackage[margin=1.15in]{geometry}
\usepackage{float}
\usepackage{amsmath}
\usepackage{amsbsy}
\usepackage{setspace}
\usepackage{amssymb}
\usepackage{esint}
\usepackage{cite}
\newfloat{algorithm}{tbp}{loa}
\floatname{algorithm}{Algorithm}
\usepackage{algorithmic}

\newlength\myindent
\makeatletter

\floatstyle{ruled}
\newfloat{algorithm}{tbp}{loa}
\floatname{algorithm}{Algorithm}

\usepackage{babel}

\begin{document}

\title{High-Dimensional Vector Semantics}

\author{M. Andrecut}

\date{January 5, 2018}

\maketitle
{

\centering Calgary, Alberta, T3G 5Y8, Canada

\centering mircea.andrecut@gmail.com

} 
\bigskip 
\begin{abstract}
In this paper we explore the "vector semantics" problem from the perspective of "almost 
orthogonal" property of high-dimensional random vectors. We show that this intriguing  
property can be used to "memorize" random vectors by simply adding them, and we provide an efficient probabilistic solution 
to the set membership problem. 
Also, we discuss several applications to word context vector embeddings, document sentences similarity, and spam filtering. 

\end{abstract}

\section{Introduction}

In many natural language processing tasks the words and the documents are represented using the "bag of words" model. 
In such a model, a document is represented by a high-dimensional vector, with the components corresponding 
to the frequency of a particular word in the document (for a detailed discussion see \cite{key-1,key-2,key-3} and the references within). 
For example, assuming an English vocabulary of $25,000$ words, 
each document will be represented by a $25,000$ dimensional vector, where the component $i$ is the frequency of the $i$th 
word in the document. The vector representation is particularly useful in text classification tasks, where 
the similarity of two documents can be simply estimated using the dot product between the vectors. 
If the vectors are normalized, then their dot product is equal to the cosine of the angle between the vectors, 
and therefore the more parallel the vectors are, the more similar the documents are. 

Another frequently encountered problem is the word vector embedding. 
In such a problem, the words are represented by high-dimensional vectors, and their "meaning" is computed from their context,  
which is modeled using the distribution of words around them (for a detailed discussion see \cite{key-1,key-2,key-3} and the references within). 
Several computational methods based on pointwise mutual information, (deep) neural networks, 
matrix factorizations or agglomerative clustering have been developed to compute the "meaning" of words. 
These efforts have culminated in identifying the words that share semantic $(dog, cat, cow)$
or syntactic $(emptied, carried, danced)$ properties, or in solving more complex problems like estimating 
the similarity between pairs of words \cite{key-3,key-4}. For example from the pairs $(king,queen)$ and $(man,woman)$, 
one can roughly recover $queen \approx king - man + woman$, by simply using linear vector algebra \cite{key-4}. 

A different approach to these problems is based on the random indexing method \cite{key-5,key-6}. 
In this approach a $d$-dimensional sparse random vector called a random label is assigned to each different word in the
text data. These labels have a small number of randomly distributed -1s and +1s, with the rest set to 0. 
In the next step, for any given word the labels for the words in its context window are added to its context vector. 
This approach is motivated by an earlier observation, that in high-dimensional spaces there are 
many more "almost orthogonal" directions than the dimensionality of the space \cite{key-7}.

Inspired by these ideas, here we explore the "vector semantics" problem from the perspective of "almost orthogonal" 
property of high-dimensional random vectors. More exactly, we extend the theoretical justification of this method by 
providing a probabilistic solution to the set membership problem (bag of words), and we discuss several potential applications to word 
context vector embeddings, document sentences similarity, and spam filtering. 
Contrary to the "expensive" machine learning methods, this method is very simple and it does not even require a "learning" process, 
however it exhibits similar properties. 

\section{Almost orthogonal random vectors}

Let us consider the set $B^d$ of random $d-$dimensional unit vectors:
\begin{equation}
\gamma = \frac{1}{\sqrt{d}}[\gamma_1,...,\gamma_d]^T, \quad \Vert \gamma \Vert =1,
\end{equation}
with the components corresponding to independent Bernoulli variables, $\gamma_i \in \lbrace -1,1\rbrace$, $i=1,...,d$, with the probability $p=1/2$. 

Let us assume that $\xi,\zeta \in B^d $ are two random vectors from $B^d$. 
These two vectors are orthogonal $\xi \perp \gamma$ if their dot product: 
\begin{equation}
\zeta^T \xi = \frac{1}{d} \sum_{i=1}^d \zeta_i \xi_i.
\end{equation}
is equal to zero. 
For high-dimensional vectors extracted from $B^d$, the expectation value of the dot product is obviously:
\begin{equation}
E(\zeta^T \xi) = \frac{1}{d} E \left( \sum_{i=1}^d \zeta_i \xi_i \right) = 0, 
\end{equation}
and the variance is:
\begin{equation}
\sigma^2 = \frac{1}{d} E\left[ \left( \sum_{i=1}^d \zeta_i \xi_i \right)^2 \right]  
= \frac{1}{d} \sum_{i=1}^d \sum_{j=1}^d \zeta_i \zeta_j E[ \xi_i \xi_j ] 
= \frac{1}{d} \sum_{i=1}^d \zeta_i^2 = \frac{1}{d}.
\end{equation}
Also, using the Chernoff bound\cite{key-8} we obtain:
\begin{equation}
\text{Pr}(\vert \zeta^T \xi \vert > \delta) < \exp \left[  -\left( \frac{\delta}{\sigma} \right)^2 \right]  = \exp \left( -d \delta^2 \right). 
\end{equation}
Thus, the probability that the two random vectors $\zeta, \xi \in B^d$ are "almost orthogonal" is given by:
\begin{equation}
\text{Pr}_{\delta}(\zeta \perp \xi) > 1 - \exp \left( -d \delta^2 \right).
\end{equation}
This means that for a relatively large dimensionality $d$, the probability that two random vectors from $B^d$ are "almost orthogonal" 
is quite high. 
For example, if $\delta = 0.05$ and $d=1,200$ we have: $\text{Pr}_{0.05}(\zeta \perp \xi) > 0.95$. 
In general, one can show that in a high dimensional space there is an exponentially large number of "almost orthogonal" randomly chosen vectors \cite{key-7,key-9}. 
Following the random indexing approach, in the next section we show that this "intriguing" property can be used to "memorize" random vectors from $B^d$ by simply adding them.

\section{Set membership problem}

Let us now consider the following set membership problem: given a set of $k$ random vectors, 
$\Xi=\lbrace \xi_1,\xi_2,...,\xi_k \mid \xi_i \in B^d, i=1,...,k \rbrace$ 
and a new random vector $\gamma \in B^d$, we want to check if $\gamma \in \Xi$. This is a typical binary decision problem, with the answer TRUE or FALSE. 
Normally, the solution requires the calculation of the dot product of $\gamma$ with each vector $\xi_i \in \Xi$, $i=1,2,...,K$. If $\exists \xi_i \in \Xi$ such that $\gamma^T \xi_i = 1$ 
then the answer is TRUE, otherwise the answer is FALSE. Thus, the solution to the set membership problem is practically a binary classifier, 
which also acts as a "set filter" in $B^d$. 

The above set membership problem can be also reformulated as a "query" problem, by asking to return the vector $\xi_{i^*} \in \Xi$,  
which is most similar to the "query" vector $\gamma$. 
In this case the answer is obtained by taking the $k$ dot products, and searching for the index $i^*$ of the product with the highest value:
\begin{equation}
i^*=\text{arg} \max_{i=1,...,k} \xi_i^T \gamma.
\end{equation}

We can see that statistically the set membership problem requires an average of $k/2$ operations (dot products) in order to provide a correct answer, 
while the "query" problem requires $k$ operations (dot products) and a sorting procedure. However, here we will show that probabilistically the 
set membership problem can be solved using only one operation (dot product). 

Let us consider the sum of all random vectors from the given set $\Xi$:
\begin{equation}
\zeta = \sum_{i=1}^k \xi_i,
\end{equation}
and the dot product of $\zeta$ with the "query" vector $\gamma$:
\begin{equation}
\eta = \gamma^T \zeta.
\end{equation}

Let us first assume that $\exists i$ such that $\xi_i = \gamma$. Then the expectation of the dot product is:
\begin{equation}
E(\eta) = E\left( \gamma^T \xi_i + \sum_{j\neq i} \gamma^T \xi_j \right)  = 1 + E\left( \sum_{j\neq i} \gamma^T \xi_j \right)  = 1 + \sum_{j\neq i} E(\gamma^T \xi_j) = 1,
\end{equation}
since $\gamma$ and $\xi_{j\neq i}$ are random vectors from $B^d$, and $E(\gamma^T \xi_{j \neq i}) = 0$ according to the previous result. 

Obviously, if $\nexists i$ such that $\xi_i = \gamma$, then the expectation of the dot product is:
\begin{equation}
E(\eta) = E\left(\sum_{j=1}^k \gamma^T \xi_j \right)  = \sum_{j=1}^k E(\gamma^T \xi_j) = 0.
\end{equation}
Also, one can easily see that the variance of the dot product is:
\begin{equation}
\sigma^2 = E\left[ \left( \sum_{j=1}^k \gamma^T \xi_j \right)^2  \right] = k/d.
\end{equation}

In order to illustrate numerically this result we consider $d=10^4$ and $k=10^3$, and we plot the value of the dot product $\eta$ in 
$10^3$ cases where the "query" vector $\gamma$ is a member, and 
respectively a non-member, of the set $\Xi$. The results are shown in Figure 1. Here we have also included the distributions of 
$\eta$ for the distinct member and non-member situations. 

\begin{figure}[t!]
\centering \includegraphics[width=10cm]{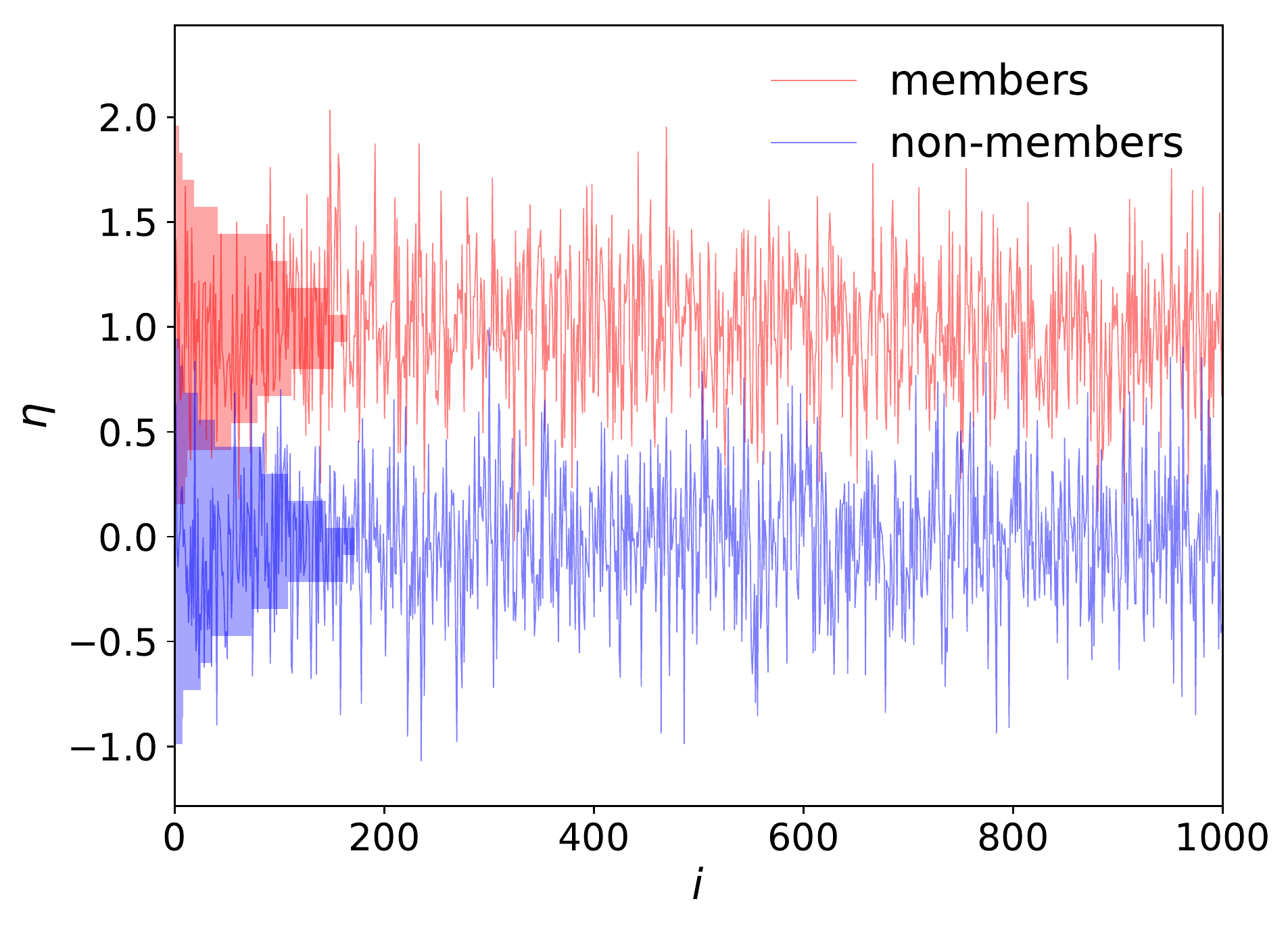}
\caption{Probabilistic solutions for the set membership problem.}
\end{figure}

\begin{figure}[t!]
\centering \includegraphics[width=10cm]{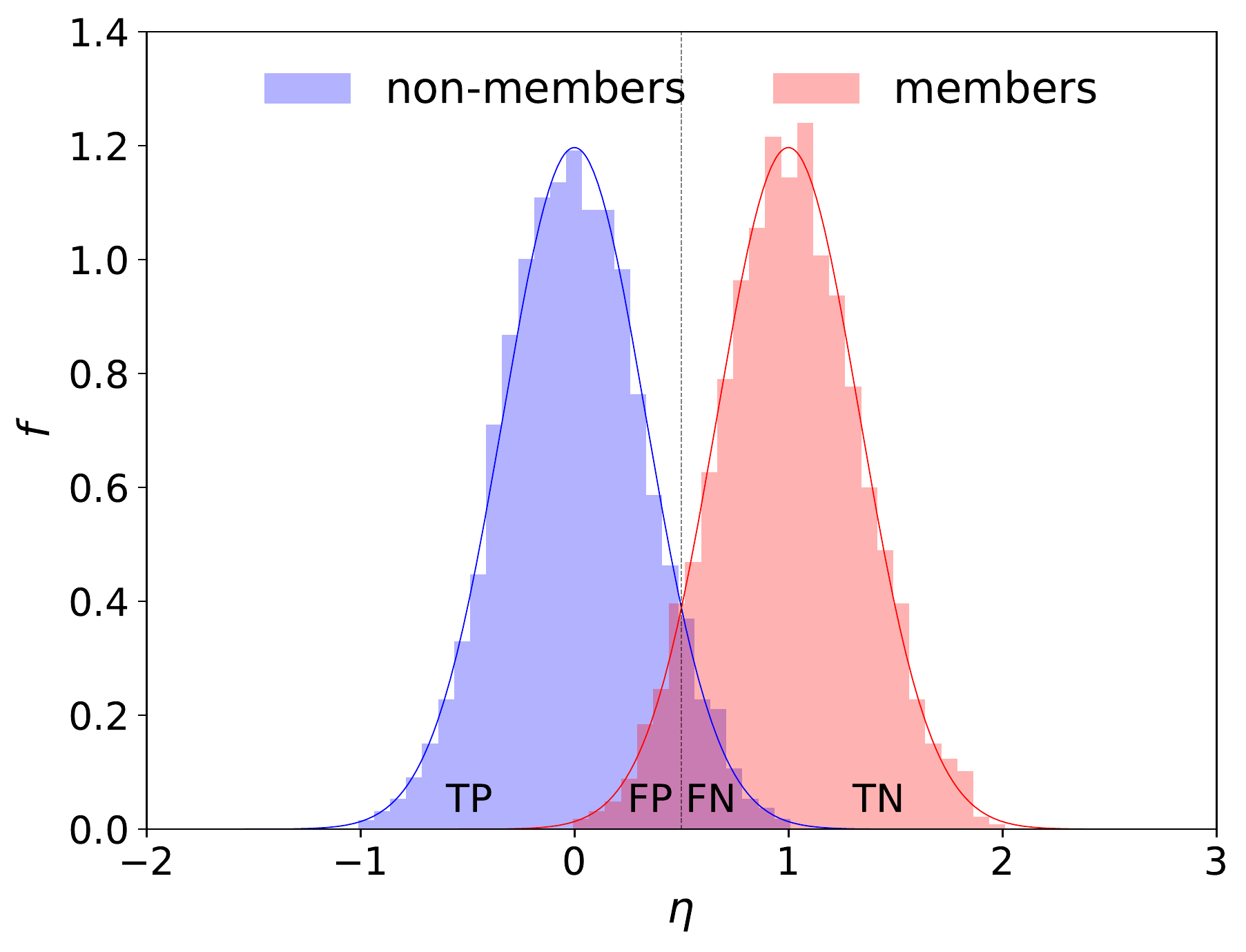}
\caption{Distributions overlap for the probabilistic set membership problem ($\sigma=1/3$).}
\end{figure}

Since the components of the vectors are Bernoulli distributed, the distribution of the $\eta$ values is binomial (scaled by a multiplying constant $1/\sqrt{d}$). 
One can also approximate the binomial distribution with a normal distribution with the mean $\mu=1$ for members, and respectively $\mu=0$ 
for non-members, both having a standard deviation: $\sigma= \sqrt{k/d}$. Therefore, we have two normal distributions $\mathcal{N}(1,\sqrt{k/d})$ 
and respectively $\mathcal{N}(0,\sqrt{k/d})$. One can estimate the classification precision and recall by using the overlap 
of these two normal distributions as a function of the standard deviation $\sigma=\sqrt{k/d}$ (as shown in Figure 2). 
Since the intersection point of these two distributions is at $\eta^*=1/2$, the overlap will be:
\begin{equation}
s(\sigma) = \text{P}(\eta > 1/2) + \text{P}(\eta < 1/2) = 1 - \Phi\left( \frac{1}{2\sigma} \right)  + \Phi\left( -\frac{1}{2\sigma} \right),
\end{equation}
where $\Phi(x)$ is the cumulative distribution function:
\begin{equation}
\Phi(x) = \frac{1}{2\pi} \int_{-\infty}^{x} \exp(-t^2/2) dt.
\end{equation} 
In fact, the overlap $s(\sigma)$ is an estimation of the sum of false positive ($FP$) and false negative ($FN$) classification cases:
\begin{equation}
s(\sigma) = FP(\sigma) + FN(\sigma),
\end{equation}
also because of the perfect symmetry of the intersecting distributions we have:
\begin{equation}
FP(\sigma) = FN(\sigma) = \frac{1}{2} s(\sigma),
\end{equation}
and respectively:
\begin{equation}
TP(\sigma) = TN(\sigma) = 1 - s(\sigma),
\end{equation}
where $TP$ and $TN$ are the true positive, and respectively true negative classification cases. 
Therefore, the classification precision and recall are equal to the following quantity:
\begin{equation}
\rho(\sigma) = \frac{TP(\sigma)}{TP(\sigma)+FP(\sigma)} = \frac{TP(\sigma)}{TP(\sigma)+FN(\sigma)} = 1- \frac{s(\sigma)}{2-s(\sigma)}.
\end{equation}

In order to illustrate numerically this result we consider $d=10^3$	and we let $k=2,...,d$. For each $k$ we compute the 
distributions from $T=10^3$ samples with $10^3$ cases where the "query" vector $\gamma$ is a member, and 
respectively a non-member, of the set $\Xi$. The obtained results for $\rho(\sigma)$ are shown in Figure 3, and they 
are in perfect agreement with the analytical estimation (13)-(18). 
It is interesting to see that for $\sigma \in (0,0.215]$ the precision (recall) is $\rho(\sigma)\geq 0.99$, and for 
$\sigma \in (0,0.375]$ we have $\rho(\sigma)\geq 0.90$. This means that for a large dimension $d$ one can solve 
the set membership problem with high probability for relatively large sets $\Xi$. 
For example if $d=1000$ one can solve with high probability ($p > 0.9$) set membership problems with up to $k = 375$ members,  
which is quite impressive, considering the simplicity of the method.

\begin{figure}[t!]
\centering \includegraphics[width=10cm]{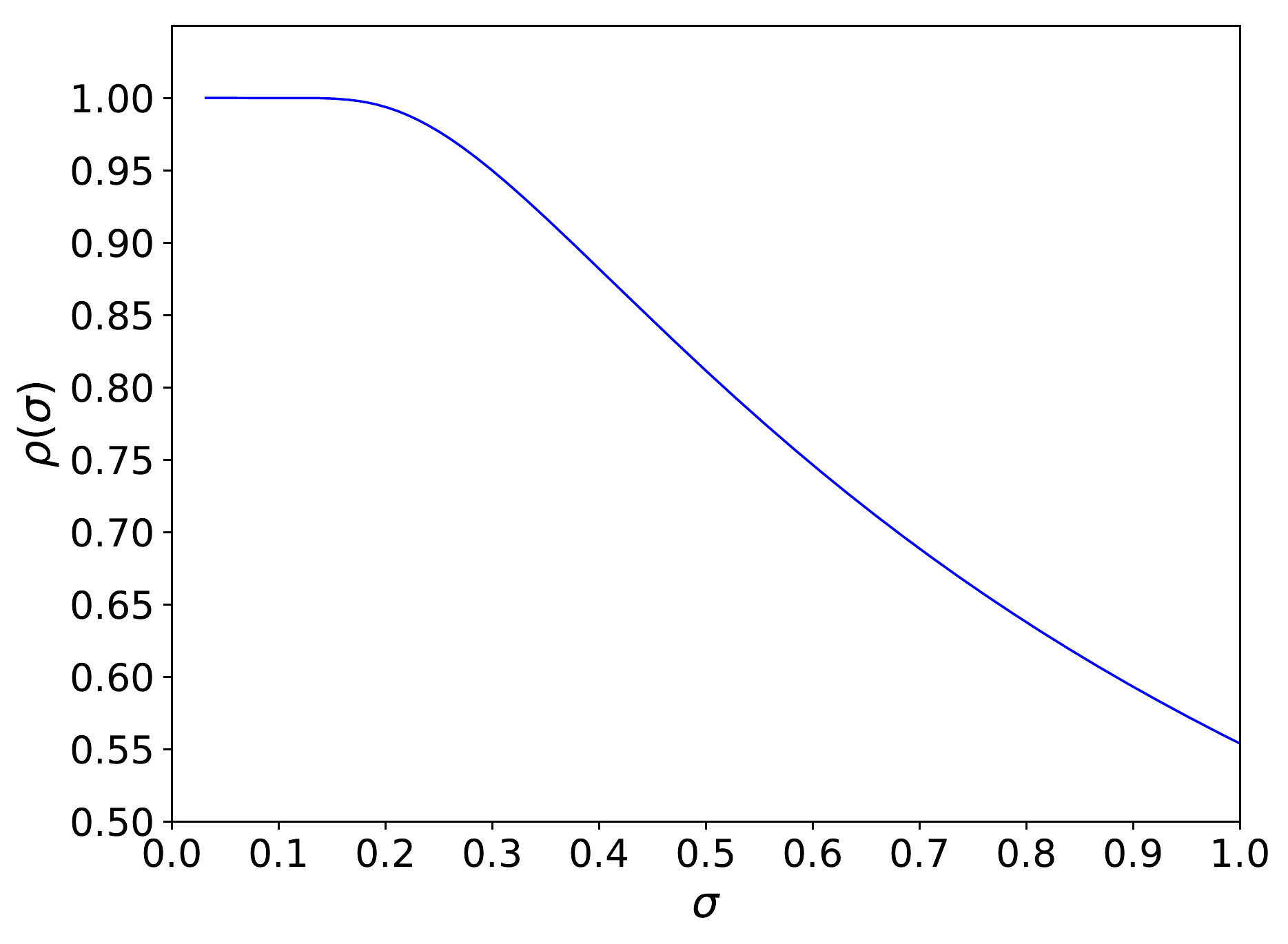}
\caption{Precision and recall $\rho(\sigma)$ for the probabilistic set membership problem.}
\end{figure}

\section{Application to natural language processing tasks}

\subsection{Word-context vectors}

We consider a vocabulary $V$ of $n$ unique words. With each word $w\in V$ we associate a randomly drawn vector 
$\xi_w$ from $B^d$. Thus, we associate the vocabulary $V= \lbrace w_i \mid i=1,...,n \rbrace$ with a set $\Xi = \lbrace \xi_i \mid \xi_i \in B^d, i=1,...,n \rbrace$
of "almost orthogonal" random vectors. 

Let us now consider a document $D$, containing $\vert D \vert$ words from the vocabulary $V$. 
For each word $w(\ell)$ in $D$ with the index $i$ in the vocabulary $V$, 
$w(\ell)\equiv w_i$, we also consider the context window of length $2L$: 
\begin{equation}
W^\ell = \lbrace w(\ell-L),...,w(\ell-1),w(\ell+1),...,w(\ell+L) \mid \ell = 1,...,\vert D \vert \rbrace.
\end{equation}
We define the context $\gamma_i$ of the word $w_i \in V$ as the sum of the word vectors from all the corresponding context windows extracted from $D$:
\begin{equation}
\gamma_i = \sum_{w(\ell) \in D} \delta(\chi(w(\ell)),i) \sum_{w(\ell+j)\in W^\ell} \xi_{\chi(w(\ell+j))},
\end{equation}
where $\chi(w)$ is the function that returns the index of the word $w$ in the vocabulary $V$, and $\delta$ is Kronecker delta function:
\begin{equation}
\delta(i,j) = 
  \begin{cases}
   1  & i=j \\
   0 & i\neq j
  \end{cases}.
\end{equation}
Thus, given a document $D$, for each word $w_i$ we calculate its context as a sum $\gamma_i$ over all windows  
at the positions $\ell$ where $w_i$ appears in the document $D$. Since, the vectors $\xi_i \in \Xi$ are "almost orthogonal", 
and according to the previously obtained result for the "set membership problem", 
the context sum can accommodate quite a large number of vectors until its "set filtering" properties will significantly deteriorate. 
For example, let's assume that the context of the word $w_i$ is the set $\Gamma_i=\lbrace w_{i_1},...,w_{i_m} \rbrace$ where 
each word $w_{i_j}$ appears a number of $\theta_{i_j}$ times. Then, the context vector of the word $w_i$ is:
\begin{equation}
\gamma_i = \sum_{j=1}^{m_i} \theta_{i_j} \xi_{i_j}.
\end{equation}
One can easily check if a word $w_{i_k}$ (associated with the vector $\xi_{i_k}$) is a "member" of the context $\gamma_i$ 
by simply taking the dot product:
\begin{equation}
\eta_{i_k} = \xi_{i_k}^T \gamma_i 
\end{equation}
which has the expectation:
\begin{equation}
E(\eta_{i_k}) = E\left( \sum_{j} \theta_{i_j} \xi_{i_k}^T \xi_{i_j} \right) 
= \theta_{i_k} + E\left( \sum_{j\neq k} \theta_{i_j} \xi_{i_k}^T \xi_{i_j} \right) 
= \theta_{i_k} \geq 1.
\end{equation}
Consequently, the variance is:
\begin{equation}
\sigma^2 = \frac{1}{d} \sum_j \theta_{i_j}.
\end{equation}
Therefore, if $\sigma \in (0,0.375]$ we have a good precision and recall.

Now let's assume that we have two words $w_i$ and $w_j$ with the context vectors $\gamma_i$ and respectively $\gamma_j$:
\begin{align}
\gamma_i &= \sum_{j=1}^{m_i} \theta_{i_j} \xi_{i_j}, \\
\gamma_j &= \sum_{k=1}^{m_j} \theta_{j_k} \xi_{j_k}.
\end{align}
One can check their similarity by taking their dot product:
\begin{equation}
\eta_{ij} = \frac{\gamma_i^T \gamma_j}{\Vert \gamma_i \Vert \Vert \gamma_j \Vert}  \in [-1,1].
\end{equation}
The expectation of $\eta_{ij}$ is:
\begin{equation}
E(\eta_{ij}) = \frac{1}{\Vert \gamma_i \Vert \Vert \gamma_j \Vert} \sum_j \sum_k \theta_{i_j}\theta_{j_k} E(\xi_{i_j}^T \xi_{j_k}),
\end{equation}
where 
\begin{equation}
E(\xi_{i_j}^T \xi_{j_k}) = 
  \begin{cases}
   1  & i_j=j_k \\
   0 & i_j\neq j_k
  \end{cases}.
\end{equation}
Thus, the similarity between $w_i$ and $w_j$ is determined by the number of identical words present in their context vectors $\gamma_i$ and $\gamma_j$. 

Let us consider an example by using the book The Adventures of Sherlock Holmes by Sir Arthur Conan Doyle, which can be downloaded from the 
Gutenberg Project \cite{key-10}. 
The document is processed by removing all the "stop words", which do not bring meaningful information to contexts (words with very high frequency like: "the"), 
we also eliminate all the non-alphanumeric words, and the remaining words are lemmatized. 
The resulted corpus has 38,812 words, with a vocabulary of 5,829 unique words. 
In Figure 4 we show the (ordered) distribution of the number of words, and of the unique number of words, in the context of each word from the resulted vocabulary, 
for a window length $2L=10$. 
One can see that very few words (176 out of 5,829, or $3\%$) have a context with a number of words larger than 375. Thus, by 
using random vectors with dimensionality $d=1000$, we can still have a quite high precision (recall) in most cases. 
Here are some interesting similarities derived from the resulted word-context vectors:
(addicted : college, theological); 
(administration : affairs, secretary); 
(advise : watson); 
(americans : finns, germans); 
(answer : say); 
(arizona : montana);
(arm : hand);
(arrange : extract);
(artery : roadway, traffic). 
Here are also some relevant examples of word-context vector operations: 
accent - german $\simeq$ proficient; 
acid - pungent $\simeq$ hydrochloric;
aged - grizzle $\simeq$ middle.

\begin{figure}[t!]
\centering \includegraphics[width=10cm]{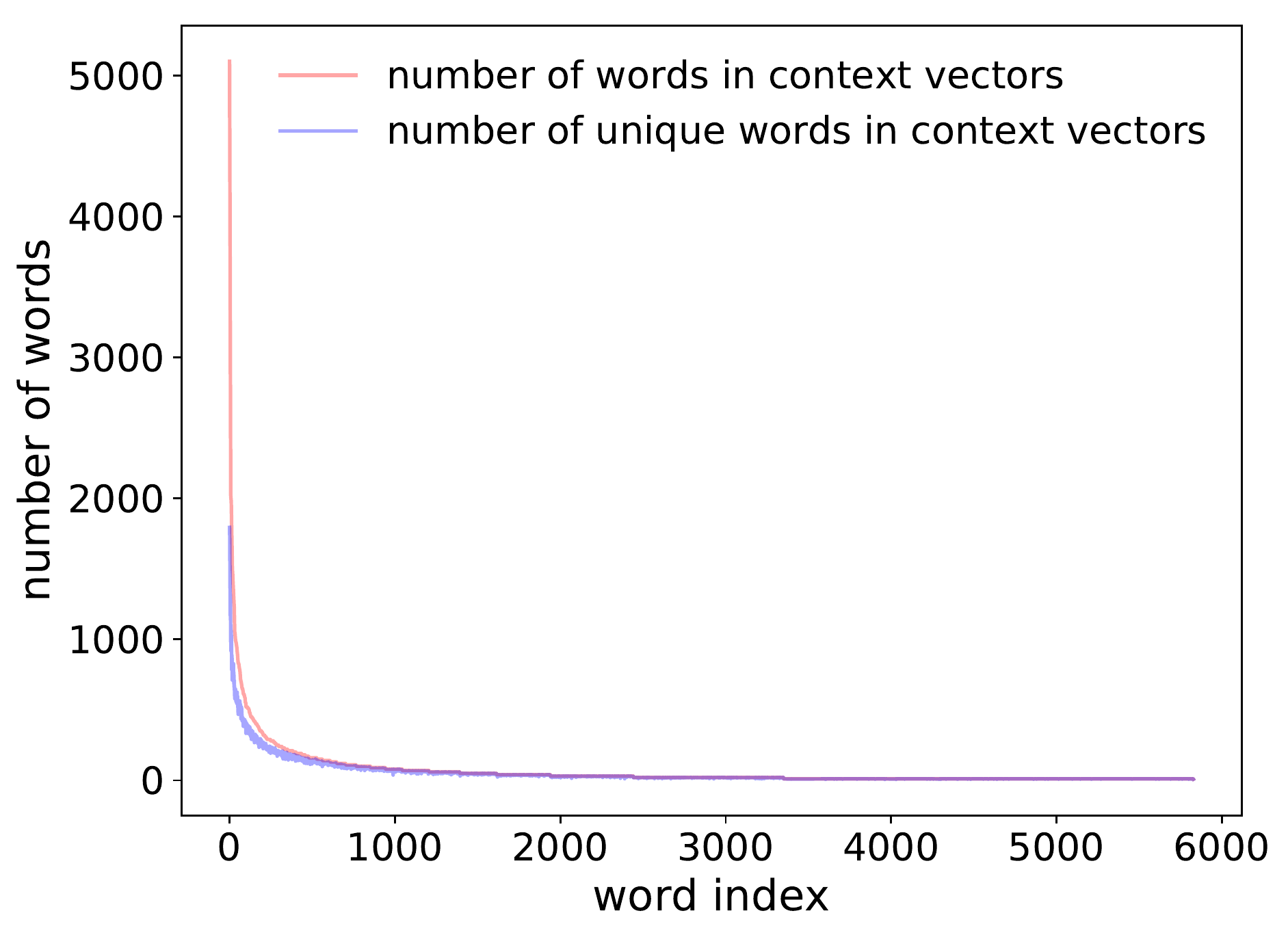}
\caption{Distribution of the number of words, and of the unique number of words, in the context of each word from the vocabulary in 
The Adventures of Sherlock Holmes.}
\end{figure}

\subsection{Sentence similarity}

Another possible application is to search a document in order to find similar sentences to a given "query" sentence. 
Again, we associate the vocabulary $V= \lbrace w_i \mid i=1,...,n \rbrace$ with a set $\Xi = \lbrace \xi_i \mid \xi_i \in B^d, i=1,...,n \rbrace$
of "almost orthogonal" random vectors, and we assume that each sentence from a document $D$ is represented by the sum of its words. Thus, the $i$th 
sentence $s_i=\lbrace w_{i1},...,w_{im_i} \rbrace$ of a document $D$, will be represented by the vector:
\begin{equation}
\gamma_i = \sum_{j=1}^{m_i} \xi_{i_j},
\end{equation}
where $i_j \equiv \chi(w_{ij})$. 

Let us also assume that the vector corresponding to the sentence query $q = \lbrace w_{1},...,w_{k} \rbrace$ is:
\begin{equation}
\zeta = \sum_{\ell=1}^{k} \xi_{\ell},
\end{equation}
where $\ell \equiv \chi(w_{\ell})$. Finding the most similar sentence to $q$ is equivalent to solving:
\begin{equation}
i^* = \text{arg} \max_i \frac{\zeta^T \gamma_i}{\Vert \zeta \Vert \Vert \gamma_i \Vert}.
\end{equation}
The expectation of the dot product $\eta_i=\nu \zeta^T \gamma_i$ 
is $E(\eta_i)=\nu n_i$, where $n_i$ is the number of words the sentences $q$ and $s_i$ have in common, and $\nu = (\Vert \zeta \Vert \Vert \gamma_i \Vert)^{-1}$. 
Also, the variance of the dot product is: $\sigma^2_i = \nu n_i/d$. Thus, with a reasonable high dimensionality $d$ we can obtain a very good precision (recall). 

For example, let's consider again the book The Adventures of Sherlock Holmes, and we ask the following naive question: 
"Who is the woman Irene in the photograph, and what is her special connection to Sherlock?"
The question is naive because after processing the "query" sentence, only the following words actually are used in the search process: 
woman, irene, photograph, special, connection, sherlock, and they are also independent of eachother, since only their "sum" is used.  

The first three sentences returned using the method described above are:
(1) "And when he speaks of Irene Adler, or when he refers to her photograph, it is always under the honourable title of the woman."; 
(2) "And yet there was but one woman to him, and that woman was the late Irene Adler, of dubious and questionable memory."; 
(3) "The photograph was of Irene Adler herself in evening dress, the letter was superscribed to Sherlock Holmes, Esq."

We also used the dot product without normalization $\eta_i= \zeta^T \gamma_i$, and the results for the first three  
returned sentences are also interesting:
(1) "To Sherlock Holmes she is always THE woman."; 
(2) "And when he speaks of Irene Adler, or when he refers to her photograph, it is always under the honourable title of the woman."; 
(3) "And yet there was but one woman to him, and that woman was the late Irene Adler, of dubious and questionable memory."

Therefore, the normalization of the sentence vectors before taking the dot product may or may not be necessary, and both cases may return relevant results.

\subsection{Spam filtering}

Spam filters are built in order to protect email users from spam and phishing messages. 
Most spam filters are word-based filters, which simply block any email that contains certain words or phrases. Another approach is based on machine learning 
techniques such as Bayesian classifiers, which must be trained on large sets of already classified spam and non-spam messages. 
Here we discuss a different approach, based on the "almost orthogonal" property of random vectors. 

As in the previously described applications, we associate the vocabulary $V= \lbrace w_i \mid i=1,...,n \rbrace$ with a set 
$\Xi = \lbrace \xi_i \mid \xi_i \in B^d, i=1,...,n \rbrace$ of "almost orthogonal" random vectors, 
and we assume that each message is represented by the sum of its words, and equivalently the sum of 
"almost orthogonal" vectors representing the words. 
Therefore, we assume that we have $m$ messages $g_j$, $j=1,...,m$, already classified, such that the associated vectors are:
\begin{equation}
\gamma_j = \sum_{i=1}^{m_j} \xi_{j_i},
\end{equation}
where $m_j$ is the number of words in the message $g_j$.
Also, the class of each message $g_j$, $j=1,...,m$, is known:
\begin{equation}
\text{class}(g_j) = 
  \begin{cases}
   1  & \text{if} \: \: g_{j} \: \: \text{spam} \\
   0  & \text{if} \: \: g_{j} \: \: \text{non-spam}
  \end{cases}.
\end{equation}
Now, let us assume that $h$ is a new message, with the associated vector:
\begin{equation}
\zeta = \sum_{\ell=1}^{k} \xi_{\ell},
\end{equation}
where $k$ is the number of words in $h$.

In order to classify $h$ as spam or non-spam we simply compute:
\begin{equation}
j^* = \text{arg} \max_{j=1,...,m} \frac{\zeta^T \gamma_j}{\Vert \zeta \Vert \Vert \gamma_j \Vert}, 
\end{equation}
and we assign to $h$ the class of $g_{j^*}$:
\begin{equation}
\text{class}(h) = \text{class}(g_{j^*})
\end{equation}
Thus, the class attributed to $h$ is the class of the most similar, and already classified message $g_{j^*}$. 

\begin{figure}[t!]
\centering \includegraphics[width=10cm]{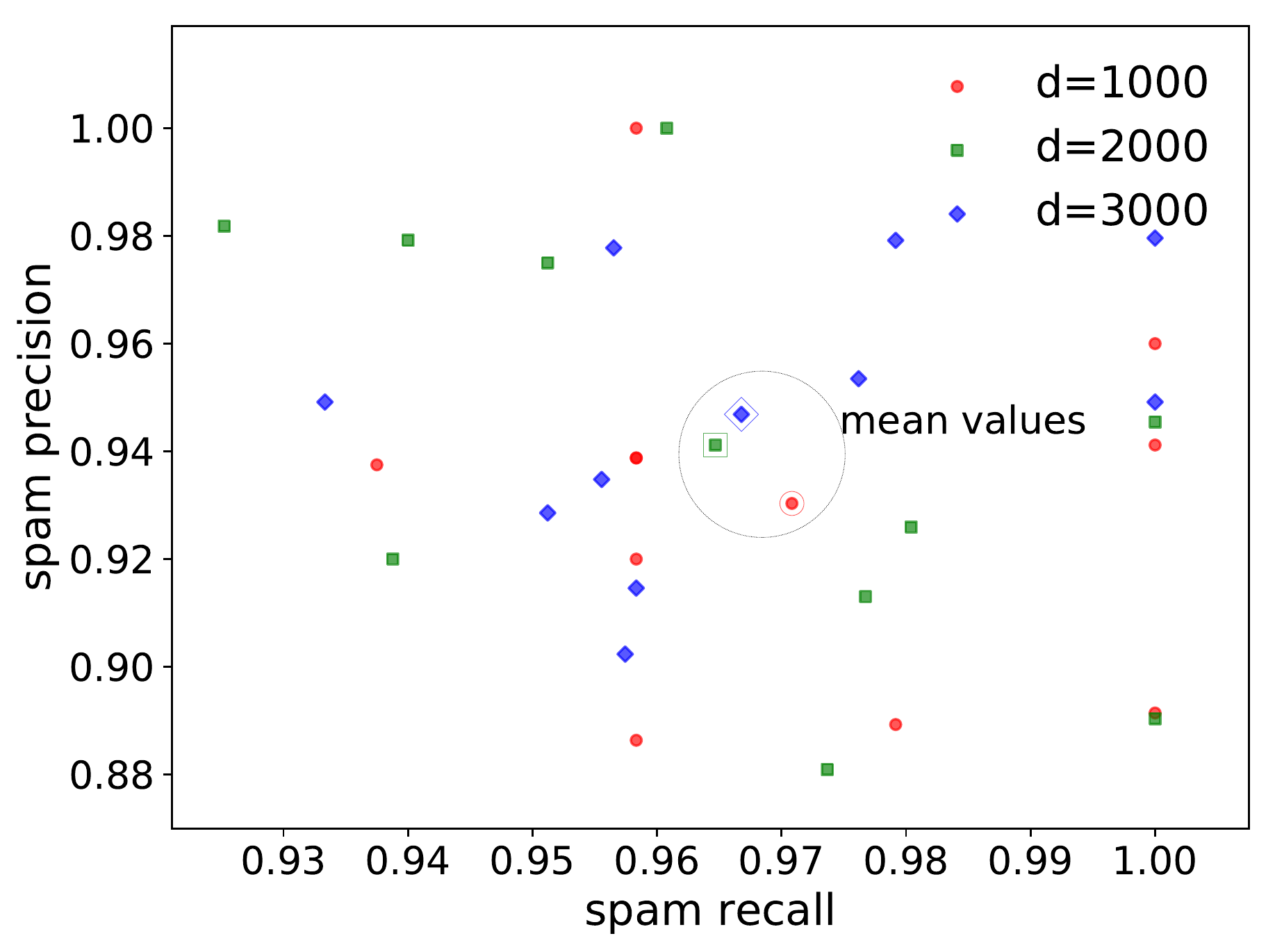}
\caption{Spam precision and recall for $d=1000,2000,3000$.}
\end{figure}

In order to evaluate this very simple method we use the Ling-Spam corpus \cite{key-11}, as described in the paper Ref. \cite{key-12}. 
The data set contains four subdirectories, corresponding to four versions of the corpus:
(1) bare: lemmatiser disabled, stop-list disabled; (2) lemm: lemmatiser enabled, stop-list disabled; 
(3) lemm-stop: lemmatiser enabled, stop-list enabled; 
(4) stop: lemmatiser disabled, stop-list enabled.

In our experiment we used the files from the first subdirectory: "bare: lemmatiser disabled, stop-list disabled". 
This directory contains 10 subdirectories (part1,..., part10), corresponding to the 10 partitions of the corpus 
used in the 10-fold cross validation experiment. In each repetition, one 
part is reserved for testing and the other 9 are used for training. 
Each one of the 10 subdirectories contains both spam and legitimate 
messages. The total number of files is 2,893. Files whose names have the form "spmsg*.txt" are spam messages. All other files are legitimate messages.

We preprocessed the messages using the spaCy Python library \cite{key-13}. 
The messages were processed by removing the "stop words" and the remaining words were lemmatized, resulting 
in a vocabulary of 54,442 unique words. 
The results for 10-fold cross validation are shown in Figure 5, for three different vector dimensionality values $d=1000,2000,3000$. 
One can see that in all three cases the average values are quite close, indicating that decreasing the dimensionality 
from 3000 to 1000 has only a slight effect on the classification precision and recall. Also, one can see that the described method 
based on "almost orthogonal" random vectors gives better results ($recall \approx 0.967, precision \approx 0.946 $ for $d=3000$) 
than the Bayesian approach described in Ref. \cite{key-12}, even though in this case there is no learning involved. 

\section{Conclusion}

In this paper we have explored a different approach to the "vector semantics" problem, which is based on the "almost 
orthogonal" property of high-dimensional random vectors. We have shown that the "almost orthogonal" 
property can be used to "memorize" random vectors by simply adding them, and we have provided an efficient probabilistic solution 
to the set membership problem. 
Also, we have discussed several applications to word and context vector embeddings, document sentences similarity, and spam filtering.
One can easily extend this approach to other problems, like for example sentiment analysis. 
Contrary to the "expensive" machine learning methods, this method is very simple and it does not even require a "learning" process, 
however it exhibits similar properties.

\end{document}